\documentclass[10pt,journal,compsoc,hidelinks]{IEEEtran}

\usepackage{hyperref}
\usepackage[utf8]{inputenc}
\usepackage[vietnamese=nohyphenation]{hyphsubst}
\usepackage[vietnamese,icelandic,english]{babel}
\usepackage{mathtools}
\usepackage[T1]{fontenc}
\usepackage{graphicx}
\usepackage{support-caption}
\usepackage{subcaption}
\usepackage{booktabs}
\usepackage{multirow}

\newcommand{\icelandic}[1]{{\selectlanguage{icelandic}#1}}

\ifCLASSOPTIONcompsoc
  \usepackage[nocompress]{cite}
\else
  \usepackage{cite}
\fi

\begin{document}
\title{An Experimental Investigation of \\ Part-Of-Speech Taggers for Vietnamese}

\author{Tuan-Phong Nguyen,
        Quoc-Tuan Truong,
        Xuan-Nam Nguyen, 
        Anh-Cuong Le
\IEEEcompsocitemizethanks{\IEEEcompsocthanksitem This work was done in 2016, when Tuan-Phong Nguyen, Quoc-Tuan Truong and Xuan-Nam Nguyen were students at the University of Engineering and Technology, VNU Hanoi, Vietnam. The corresponding author, Anh-Cuong Le, is with the Faculty of Information Technology, Ton Duc Thang University, Ho Chi Minh city, Vietnam.\protect\\
E-mail: leanhcuong@tdt.edu.vn.
}
}








\IEEEtitleabstractindextext{%
\begin{abstract}
Part-of-speech (POS) tagging plays an important role in Natural Language Processing (NLP). Its applications can be found in many NLP tasks such as named entity recognition, syntactic parsing, dependency parsing and text chunking. In the investigation conducted in this paper, we utilize the technologies of two widely-used toolkits, ClearNLP and Stanford POS Tagger, as well as develop two new POS taggers for Vietnamese, then compare them to three well-known Vietnamese taggers, namely JVnTagger, vnTagger and RDRPOSTagger. We make a systematic comparison to find out the tagger having the best performance. We also design a new feature set to measure the performance of the statistical taggers. Our new taggers built from Stanford Tagger and ClearNLP with the new feature set can outperform all other current Vietnamese taggers in term of tagging accuracy. Moreover, we also analyze the affection of some features to the performance of statistical taggers. Lastly, the experimental results also reveal that the transformation-based tagger, RDRPOSTagger, can run significantly faster than any other statistical tagger.
\end{abstract}
}

\maketitle

\IEEEdisplaynontitleabstractindextext
\IEEEpeerreviewmaketitle

\IEEEraisesectionheading{\section{Introduction}\label{sec:intro}}
In Natural Language Processing, part-of-speech (POS) tagging is the process to assign a part-of-speech to each word in a text according to its definition and context. POS tagging is a core task of NLP. The part-of-speech information can be used in many other NLP tasks, including named entity recognition, syntactic parsing, dependency parsing and text chunking. In common languages such as English and French, studies in POS tagging are very successful. Recent studies for these languages \cite{toutanova2003feature, choi2012fast,sogaard2011semisupervised, huang2015bidirectional, denis2012coupling} can yield state-of-the-art results at approximately 97-98\% for overall accuracy. However, for less common languages such as Vietnamese, current results are not as good as for Western languages. Recent studies on Vietnamese POS tagging such as \cite{le2010empirical, nguyen2014rdrpostagger} can only achieves approximately 92-93\% for precision.

Several POS tagging approaches have been studied. The most common ones are stochastic tagging, rule-based tagging and transformation-based tagging whereas the last one is a combination of the others. All of these three approaches treat POS tagging as a supervised problem that requires a pre-annotated corpus as training data set. For English and other Western languages, almost studies that provide state-of-the-art results are based on the supervised learning. Similarly, the most widely-used taggers for Vietnamese, JVnTagger \cite{nguyen2010jvntextpro}, vnTagger \cite{le2010empirical} and RDRPOSTagger \cite{nguyen2014rdrpostagger}, also treat POS tagging as a supervised learning problem. While JVnTagger and vnTagger are stochastic-based tagger, RDRPOSTagger implements a transformation-based approach. Although these three taggers are reported to have the highest accuracies for Vietnamese POS tagging, they can only give the precision of 92-93\%. Meanwhile, two well-known open-source toolkits, ClearNLP \cite{choi2012fast} and Stanford POS Tagger \cite{toutanova2003feature}, which use stochastic tagging algorithms can provide overall accuracies of over 97\% for English. It would be unfair to compare the results for two different languages because they have distinct characteristics. Therefore, our questions are \textit{``How well can the two international toolkits perform POS tagging for Vietnamese?''} and \textit{``Which is the most effective approach for Vietnamese part-of-speech tagging?''}. The purpose of the investigation conducted in this paper is to answer those questions by doing a systematic comparison of the taggers. Beside the precision of taggers, their tagging speed is also considered because many recent NLP tasks have to deal with very large-scale data in which speed plays a vital role.  

For our experiments, we use Vietnamese Treebank corpora \cite{nguyen2009building} which is the most common corpus and has been utilized by many studies on Vietnamese POS tagging and is one resource from a national project named ``Building Basic Resources and Tools for Vietnamese Language and Speech Processing'' (VLSP)\footnote{\url{https://vlsp.org.vn/}}. Vietnamese Treebank contains about 27k POS-tagged sentences. In spite of its popularity, there have been several errors in this data that can draw the precision of taggers. All of those errors that we detected are also reported in this paper. By using 10-fold cross-validation method on the configured corpus, it is revealed that the new taggers we built from ClearNLP and Stanford POS Tagger produce the most accurate results at 94.19\% and 94.53\% for precision, which also are the best Vietnamese POS tagging results known to us. Meanwhile, the highest tagging speed belongs to the transformation-based tagger, RDRPOSTagger, which can assign tags for over 161k words per second in average while running on a personal computer.

The remainder of this paper is organized as follows. In section~\ref{sec:background}, we briefly introduce general knowledge about the main approaches that have been applied in POS tagging task. We also give some information about particular characteristics of Vietnamese language and the experimental data, Vietnamese Treebank corpora. Section~\ref{sec:method} represents the methods used by the POS taggers. In section~\ref{sec:exp}, we talk about the main contribution of this paper including the error fixing process for the experimental data, the experimental results on the taggers and the comparison of their accuracies and tagging speeds. Finally, we conclude this paper in section~\ref{sec:con}.

\section{Background}
\label{sec:background}
This section provides some background information of part-of-speech tagging approaches that have been used so far. The related works are also covered. Moreover, we also give some details about Vietnamese language and Vietnamese Treebank. 

\subsection{Approaches for POS tagging}
Part-of-speech tagging is commonly treated as a supervised learning problem. Each POS tagger takes the information from its training data to determine the tag for each word in input text. In most cases, a word might have only one possible tag. The other case is that a word has several possible tags; or a word has not appeared in the lexicon extracted from the training data. The process to choose the right tag for a word in these cases is based on which kind of used tagging algorithm. There are three main kinds of tagging approaches within POS tagging, which are stochastic tagging, rule-based tagging and transformation-based tagging.

Stochastic (probabilistic) tagging approach is one of the most widely-used ones in recent studies for POS tagging. The general idea of stochastic taggers is that they make use of training corpus to determine the probability of a specific word having a specific tag in a given context. Common methods of stochastic approach are Maximum Entropy (MaxEnt), Conditional Random Fields (CRFs), Hidden Markov Models (HMMs). Many studies on English POS tagging using stochastic approaches can gain state-of-the-art results, such as \cite{toutanova2003feature, choi2012fast, huang2015bidirectional}.

Rule-based tagging is actually different from stochastic tagging. Rule-based tagging algorithm uses a set of hand-written rules to determine the tag for each word. This leads to a fact that this set of rules must be properly written and checked by experts on linguistic.

Meanwhile, transformation-based tagging is a combination of the features of the two algorithms above. This algorithm applies disambiguation rules like the rule-based tagging, but these rules are not hand-written. They are automatically extracted from the training corpus. Taggers using this kind of algorithm are usually referred to Brill's one \cite{brill1992simple}. There are three main steps in his algorithm. Firstly, the tagger initially assigns for each word in the input text with the tag which is the most frequent for this word in the lexicon extracted from the training corpus. After that, it traverses through a list of transformation rules to choose the rule that enhances tagging accuracy the most. Then this transformation rule will be applied to every word. The loop through three stages is continued until it optimizes the tagging accuracy.

For all of those approaches listed above, a pre-annotated corpus is prerequisite. On the other hand, there is also unsupervised POS tagging algorithm \cite{prins2001unsupervised, brill1995unsupervised} that does not require any pre-tagged corpus.

For Vietnamese POS tagging, \cite{tran2009experimental} compares three tagging methods which are CRFs-based, MEMs-based and SVM-based tagging. However, the comparison does not contain terms of unknown words accuracy and tagging speed. Moreover, all of those methods are based on stochastic tagging. It is necessary to systematically compare all of those characteristics of the taggers in a same evaluation scheme and also the accuracies of different kinds of approach to find out the most accurate one for Vietnamese POS tagging.

\selectlanguage{vietnamese}
\subsection{Vietnamese language}
In this section, we talk about some specific characteristics of Vietnamese language compared to the Western languages and also some information of Vietnamese Treebank, the corpus which we use for experiments.

\subsubsection{The language}

Vietnamese is an Austroasiatic language and the national and official language of Vietnam. It is the native language of Kinh people. Vietnamese is spoken throughout the world because of Vietnamese emigration. The Vietnamese alphabet in use today is a Latin alphabet with additional diacritics and letters.

In Vietnamese, there is no word delimiter. Spaces are used to separate the syllables rather than the words. For example, in the sentence \textit{``[học sinh] [học] [sinh học]''} (\textit{``students study biology''}), there are two times that \textit{``học sinh''} appears, the first space between \textit{``học sinh''} is the separation of two syllables of the word \textit{``học sinh''} (\textit{``students''}), however, the second one is not.

Vietnamese is an inflectionless language whose word forms never change as in occidental languages. There are many cases in that a word has more than one part-of-speech tags in different contexts. For instance, in the sentence \textit{``[học sinh] [ngồi] [quanh] [bàn]\textsubscript{1} [để] [bàn]\textsubscript{2} [về] [bài] [toán]''} (\textit{``students sit around the [table]\textsubscript{1} in order to [discuss]\textsubscript{2} about a Math exercise''}), the first word \textit{bàn} is a noun but the second one is a verb. Part-of-speech for Vietnamese words is usually ambiguous so that they must be classified based on their syntactic functions and meaning in their current context.

\subsubsection{Vietnamese Treebank}

Vietnamese Treebank \cite{nguyen2009building} is the largest annotated corpora for Vietnamese. It is one of the resources from the KC01/06-10 project named ``Building Basic Resources and Tools for Vietnamese Language and Speech Processing'' which belongs to the National Key Science and Technology Tasks for the 5-Year Period of 2006-2010. The first version of the treebank consists of 10,165 sentences which are manually segmented and POS-tagged. This number in the current version of the treebank is increased to 27,871 annotated sentences\footnote{\url{https://vlsp.org.vn/resources-vlsp2013}}. The raw texts of the treebank are collected from the social and political sections of the Youth online daily newspaper. The minimal and maximal sentence lengths are 1 and 165 words respectively.

\selectlanguage{english}
\begin{table}[t]
\footnotesize
\renewcommand{\arraystretch}{1.3}
\caption{Vietnamese tagset.}
\begin{center}
\label{tagset}
\begin{tabular}{c|c|c}
\hline
\textbf{No.} & \textbf{Category} & \textbf{Description} \\ \hline
1     & Np    & Proper noun     				\\ \hline
2     & Nc    & Classifier      				\\ \hline
3     & Nu    & Unit noun       				\\ \hline
4     & N     & Common noun     				\\ \hline
5     & V     & Verb     						\\ \hline
6     & A     & Adjective       				\\ \hline
7     & P     & Pronoun    						\\ \hline
8     & R     & Adverb   						\\ \hline
9     & L     & Determiner      				\\ \hline
10    & M     & Numeral    						\\ \hline
11    & E     & Preposition     				\\ \hline
12    & C     & Subordinating conjunction     	\\ \hline
13    & Cc    & Coordinating conjunction    	\\ \hline
14    & I     & Interjection    				\\ \hline
15    & T     & Auxiliary, modal words      	\\ \hline
16    & Y     & Abbreviation    				\\ \hline
17    & Z     & Bound morpheme  				\\ \hline
18    & X     & Unknown    						\\ \hline
\end{tabular}
\end{center}
\end{table}

The tagset designed for Vietnamese Treebank is presented in Table \ref{tagset}. Beside these eighteen basic tags, there are also compound tags such as \textit{Ny} (abbreviated noun), \textit{Nb} (foreign noun) or \textit{Vb} (foreign verb).

\selectlanguage{english}
\section{Method analysis}
\label{sec:method}
This section provides information about the general methods used by current Vietnamese POS taggers and two taggers for common languages. While RDRPOSTagger uses a transformation-based learning approach, all of four other taggers, ClearNLP, Stanford POS Tagger, vnTagger and JVnTagger, are stochastic-based taggers using either MaxEnt, CRFs models or support vector classification.

\subsection{Existing Vietnamese POS taggers}

\subsubsection{JVnTagger}
JVnTagger is a stochastic-based POS tagger for Vietnamese and is implemented in Java. This tagger is based on CRFs and MaxEnt models. JVnTagger is a branch product of VLSP project and also a module of JVnTextPro, a widely used toolkit for Vietnamese language processing developed by Nguyen and Phan \cite{nguyen2010jvntextpro}. This tagger is also called by the other name, VietTagger.

\begin{table*}[t]
\footnotesize
\renewcommand{\arraystretch}{1.3}
\caption{Default feature set used in JVnTagger. $w_{i}$: the word at position $i$ in the 5-word window. $t_i$: the POS tag of $w_{i}$.}
\begin{center}
\label{table:JVnTaggerFeats}
\begin{tabular}{c|c|l}
\hline
\textbf{Model} & \textbf{Type} & \multicolumn{1}{c}{\textbf{Template}} 												\\ \hline
\multirow{3}{*}{\begin{tabular}[c]{@{}c@{}}\\ \\ \\ \\ \\MaxEnt\\ and\\ CRFs\end{tabular}}

	& Lexicon    		& \begin{tabular}[c]{@{}l@{}}$w_{\{-2, -1, 0, 1, 2\}}$
												\\ $(w_{-1}, w_{0})$, $(w_{0}, w_{1})$\end{tabular}  				\\ \cline{2-3} 
	
	& Binary 			& \begin{tabular}[c]{@{}l@{}}$w_{i}$ contains all uppercase characters or not $(i=-1,0)$,
												\\ $w_{i}$ has the initial character uppercase or not $(i=-1,0)$,
												\\ $w_{i}$ is a number or not $(i=-1,0,1)$,
												\\ $w_{i}$ contains numbers or not $(i=-1,0,1)$,
												\\ $w_{i}$ contains hyphens or not $(i=-1,0)$,
												\\ $w_{i}$ contains commas or not $(i=-1,0)$,
												\\ $w_{i}$ is a punctuation mark or not $(i=-1,0,1)$\end{tabular} 	\\ \cline{2-3} 
	& \begin{tabular}[c]{@{}c@{}}Vietnamese\\ specialized features\end{tabular}
						& \begin{tabular}[c]{@{}l@{}}possible tags of $w_{i}$ in dictionary $(i=-1,0,1)$,
												\\ $w_{0}$ is full repetitive or not,
												\\ $w_{0}$ is partial repetitive or not,
												\\ the first syllable of $w_0$,
												\\ the last syllable of $w_0$\end{tabular}          				\\ \hline
CRFs
	& Edge feature       & $(t_{-1}, t_{0})$                                                                     	\\ \hline
\end{tabular}
\end{center}
\end{table*}

\selectlanguage{vietnamese}
There are two kinds of feature used in JVnTagger, which are context features for both CRFs and MaxEnt models and an edge feature for CRFs model as listed in Table~\ref{table:JVnTaggerFeats}. Both models of JVnTagger use 1-gram and 2-gram features for predicting tags of all words. For unknown words, this toolkit uses some rules to detect whether each word is in a specific form or not to determine its part-of-speech tag. 

Additionally, there is a particular feature extracted by looking up the current word in a tags-of-word dictionary which contains possible tags of over 31k Vietnamese words extracted before. This feature applies for both the current word, the previous and the next words. Besides, in Vietnamese, repetitive word is a special feature, therefore, JVnTagger adds full-repetitive and partial-repetitive word features to enhance the accuracy of predicting tag \textit{A} (adjective) as well. Word prefix and suffix are also vital features in POS tagging task of many other languages.

The CRFs model of JVnTagger had been trained by FlexCrfs toolkit \cite{phan2005flexcrfs}. Due to the nature of CRFs model, there is an edge feature extracted directly by FlexCrfs as described in Table~\ref{table:JVnTaggerFeats}.

The F-measure results of JVnTagger are reported at 90.40\% for CRFs model and 91.03\% for MaxEnt model using 5-fold cross-validation evaluation on Vietnamese Treebank corpus of over 10k annotated sentences.

\selectlanguage{english}

\subsubsection{vnTagger}
vnTagger~\cite{le2010empirical}
is also a stochastic-based POS tagger for Vietnamese. The main method of this tagger is Maximum Entropy. vnTagger is written in Java and its architecture is mainly based on the basis of Stanford POS Tagger \cite{toutanova2003feature}.

\begin{table*}[t]
\footnotesize
\renewcommand{\arraystretch}{1.3}
\caption{Default feature set used in vnTagger.}
\label{table:vnTaggerFeats}
\begin{center}
\begin{tabular}{c|l}
\hline
\textbf{Usage} & \multicolumn{1}{c}{\textbf{Template}} 														\\ \hline
All words      & \begin{tabular}[c]{@{}l@{}}$w_{\{-1, 0, 1\}}$
										\\ $t_{-1}, (t_{-2}, t_{-1})$\end{tabular} 							\\ \hline
Unknown words  & \begin{tabular}[c]{@{}l@{}}$w_{0}$ contains a number or not,
										\\ $w_{0}$ contains an uppercase character or not,
										\\ $w_{0}$ contains all uppercase characters or not,
										\\ $w_{0}$ contains a hyphen or not,
										\\ the first syllable of $w_0$,
										\\ the last syllable of $w_0$,
										\\ conjunction of the two first syllables of $w_0$,
										\\ conjunction of the two last syllables of $w_0$,
										\\ number of syllables in $w_{0}$\end{tabular} 						\\ \hline
\end{tabular}
\end{center}
\end{table*}

There are two kinds of feature used in the MaxEnt model of this tagger, which are presented in Table~\ref{table:vnTaggerFeats}. The first one is the set of features used for all words. This tagger uses a one-pass, left-to-right tagging algorithm, which only make use of information from history. It only captures 1-gram features for words in a window of size 3, and the information of the tags in the left side of the current words. The other kind of feature is used for predicting tags of unknown words. These features mainly help to catch the word shape.

The highest accuracy is reported at 93.40\% in overall and 80.69\% for unknown words when using 10-fold cross-validation on Vietnamese Treebank corpus of 10,165 annotated sentences.


\subsubsection{RDRPOSTagger}
RDRPOSTagger \cite{nguyen2014rdrpostagger} is a Ripple Down Rules-based Part-Of-Speech Tagger which is based upon transformation-based learning, a method which is firstly introduced by Eric Brill \cite{brill1992simple} as mentioned above. 
For English, it reaches accuracy scores of up to 96.57\% when training and testing on selected sections of the Penn WSJ Treebank corpus \cite{marcus1993building}. For Vietnamese, it approaches 93.42\% for overall accuracy using 5-fold cross-validation on Vietnamese Treebank corpus of 28k annotated sentences. This toolkit has both Java-implemented and Python-implemented versions.

The difference between the approach of RDRPOSTagger to Brill's is that RDRPOSTagger exploits a failure-driven approach to automatically restructure transformation rules in the form of a Single Classification Ripple Down Rules (SCRDR) tree. It accepts interactions between rules, but a rule only changes the outputs of some previous rules in a controlled context. All rules are structured in a SCRDR tree which allows a new exception rule to be added when the tree returns an incorrect classification.

\begin{figure}[t]
\centering
\includegraphics[width=.4\textwidth]{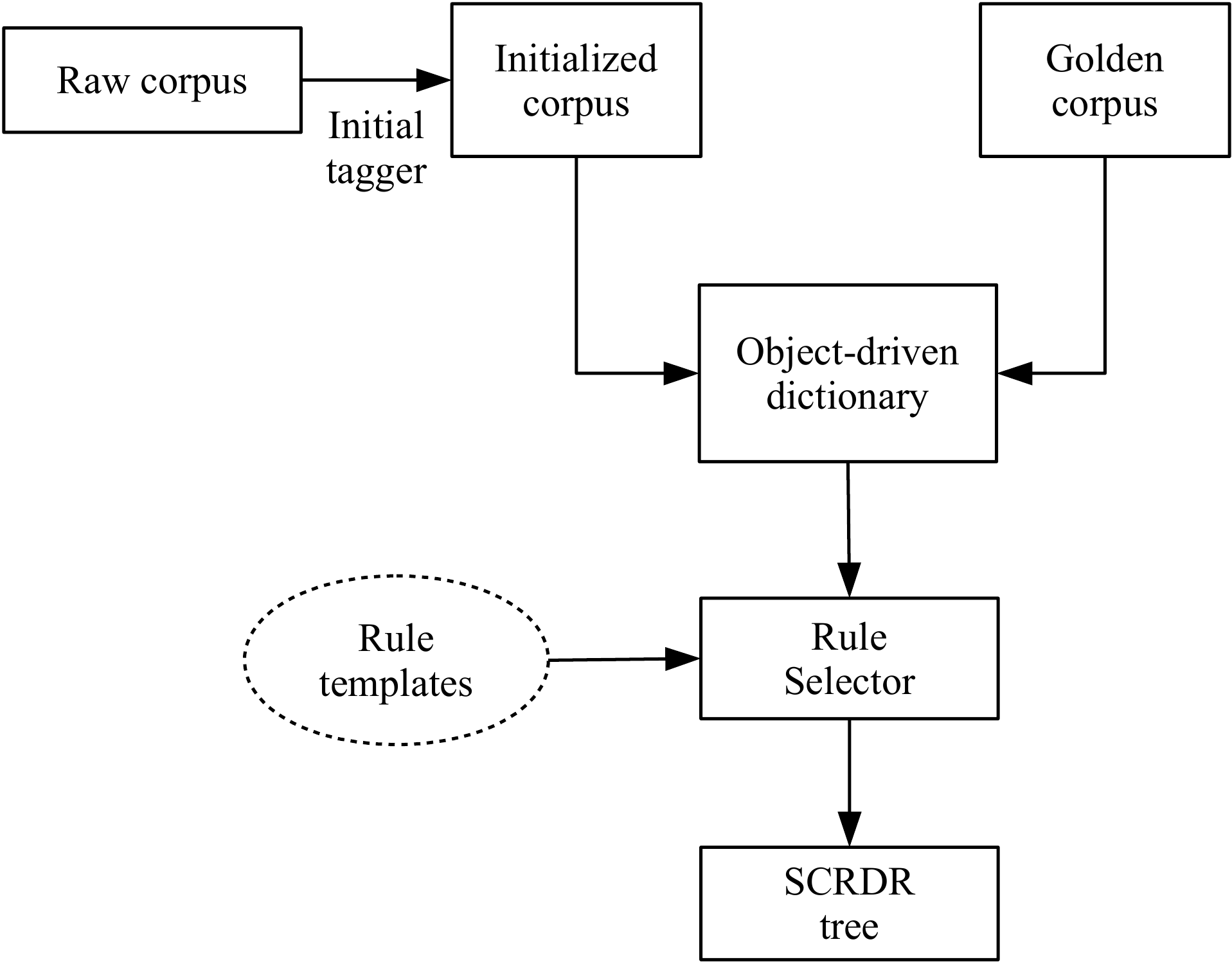}
\caption{Diagram demonstrating the learning process of the RDRPOSTagger learner.}
\label{rdrlearner}
\end{figure}

The learning process of the tagger is described in Figure~\ref{rdrlearner}. The initial tagger developed in this toolkit is based on the lexicon which is generated from the golden-standard corpus. To deal with unknown words, the initial tagger utilizes several regular expressions or heuristics whereas the most frequent tag in the training corpus is exploited to label unknown words. The initialized corpus is returned by performing the initial tagger on the raw corpus. By comparing the initialized corpus with the golden one, an object-driven dictionary of pairs ({\it Object}, {\it correctTag}) is produced in which {\it Object} captures the 5-word window context covering the current word and its tag from the initialized corpus, and the {\it correctTag} is the corresponding tag of the current word in the golden corpus.

There are 27 rule templates applied for Rule selector to select the most suitable rules to build the SCRDR tree. The templates are presented in Table~\ref{table:rdrtemplates}. The SCRDR tree of rules is initialized by building the default rule and all exception rules of the default one in the form of {\it if currentTag = ``\textbf{TAG}'' then tag = ``\textbf{TAG}''} at the layer-1 exception structure. The learner then generates new exception rules to every node of the tree due to three constraints described in \cite{nguyen2014robust}.

\begin{table}[t]
\footnotesize
\renewcommand{\arraystretch}{1.3}
\caption{Short descriptions of rule templates used for Rule selector of RDRPOSTagger.}
\label{table:rdrtemplates}
\begin{center}
\begin{tabular}{c|l|l}
\hline
\textbf{No.}   	& \multicolumn{1}{c|}{\textbf{Type}}    & \multicolumn{1}{c}{\textbf{Template}}   							\\ \hline
1 				& Word     								& $w_{\{-2, -1, 0, 1, 2\}}$      									\\ \hline
2 				& Word bigrams    						& \begin{tabular}[c]{@{}l@{}}$(w_{-2}, w_0), (w_{-1}, w_0),$
																	\\ $(w_{-1}, w_1), (w_0, w_1), (w_0, w_2)$\end{tabular}	\\ \hline
3 				& Word trigrams   						& \begin{tabular}[c]{@{}l@{}}$(w_{-2}, w_{-1}, w_0), (w_{-1}, w_0, w_1),$
																	\\ $(w_0, w_1, w_2)$\end{tabular}      					\\ \hline
4 				& POS tags   							& $t_{\{-2, -1, 0, 1, 2\}}$       									\\ \hline
5 				& POS bigrams     						& $(t_{-2}, t_{-1}), (t_{-1}, t_{1}), (t_{1}, t_{2})$    			\\ \hline
6 				& Combined   							& \begin{tabular}[c]{@{}l@{}}$(t_{-1}, w_0), (w_0, t_1), (t_{-1}, w_0, t_1),$
																	\\ $(t_{-2}, t_{-1}, w_0), (w_0, t_1, t_2)$\end{tabular}\\ \hline
7 & Suffix   	& suffixes of length 1 to 4 of $w_{0}$       																\\ \hline
\end{tabular}
\end{center}
\end{table}

The tagging process of this tagger firstly assigns tags for unlabeled text by using the initial tagger. Next, for each initially tagged word, the corresponding {\it Object} will be created. Finally, each word will be tagged by passing its object through the learned SCRDR tree. If the default node is the last fired node satisfying the object, the final tag returned is the tag produced by the initial tagger.

\subsection{POS taggers for common languages}

\subsubsection{Stanford POS Tagger}
Stanford POS Tagger \cite{toutanova2003feature} is also a Java-implemented tagger based on stochastic approach. This tagger is the implementation of a log-linear part-of-speech tagging algorithm described in \cite{toutanova2003feature} and is developed 
at Stanford University. 
At the time of this work (i.e., 2016), Stanford POS Tagger has pre-trained models for English, Chinese, Arabic, French and Germany. It can be re-trained in any other language.

The approach described in \cite{toutanova2003feature} is based on two main factors, a cyclic dependency network and the MaxEnt model. General idea of the cyclic (or bidirectional) dependency network is to overcome weaknesses of the unidirectional case. In the unidirectional case, only one direction of the tagging sequence is considered at each local point. For instance, in a left-to-right first-order HMM, the current tag $t_{0}$ is predicted based on only the previous tag $t_{-1}$ and the current word $w_{0}$. However, it is clear that the identity of a tag is also correlated with tag and word identities in both left and right sides. The approach of Stanford POS Tagger follows this idea combined with Maximum Entropy models to provide efficient bidirectional inference. 

\begin{figure}[t]
\begin{subfigure}{.5\textwidth}
  \centering
  \includegraphics[width=.7\linewidth]{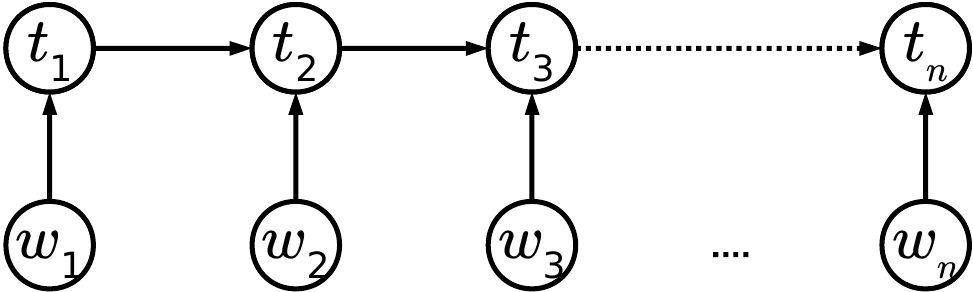}
  \caption{Left-to-Right Inference}
  \label{fig:l2r}
\end{subfigure}
\begin{subfigure}{.5\textwidth}
  \centering
  \includegraphics[width=.7\linewidth]{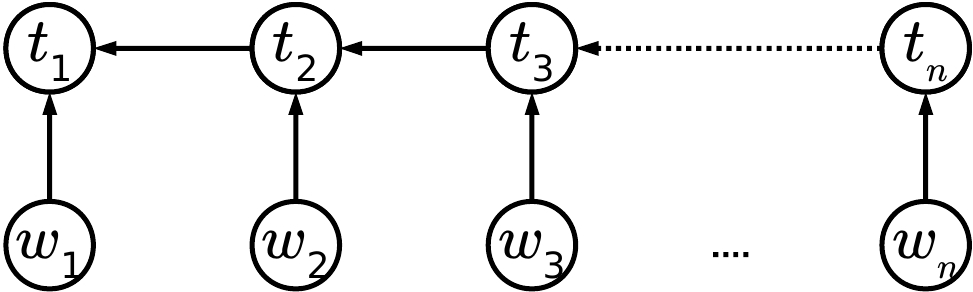}
  \caption{Right-to-Left Inference}
  \label{fig:r2l}
\end{subfigure}
\begin{subfigure}{.5\textwidth}
  \centering
  \includegraphics[width=.7\linewidth]{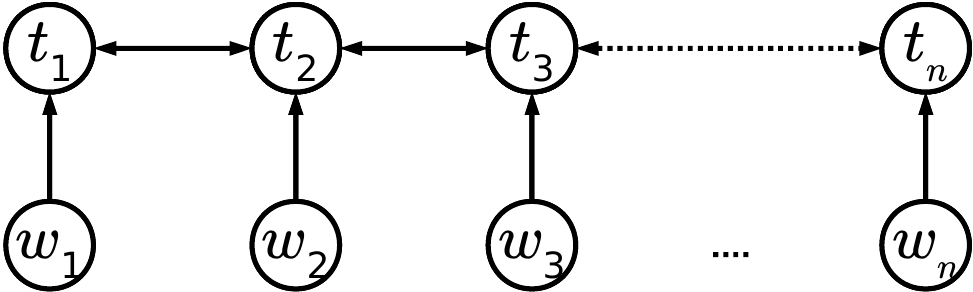}
  \caption{Bidirectional Dependency Network}
  \label{fig:bidirect}
\end{subfigure}
\caption{Dependency networks.}
\label{fig:stanford:bidirect}
\end{figure}

As reported in \cite{toutanova2003feature}, with many rich bidirectional-context features and a few additional handcrafted features for unknown words, Stanford POS Tagger can reach the overall accuracy of 97.24\% and unknown word accuracy of 89.04\%.

\subsubsection{ClearNLP}
ClearNLP \cite{choi2012fast} is a toolkit written in Java that contains low-level NPL components (e.g., dependency parsing, named entity recognition, sentiment analysis, part-of-speech tagging), developed by NLP Research Group
at Emory University. In our experiments, we use the last released version of ClearNLP – version 3.2.0.

The POS tagging component in ClearNLP is a implementation of the method described in \cite{choi2012fast}. General idea of this method is to have two models in the tagger and find the most suitable model to assign tags for input sentence based on its domain. Firstly, two separated models, one is optimized for a general domain and the other is optimized for a domain specific to the training data, are trained. They suppose that the domain-specific and generalized models perform better to sentences similar and not similar to the training data, respectively. Hence, during decoding, they dynamically select one of the models by measuring similarities between input sentences and the training data. Some first versions of ClearNLP use dynamic model selection but later versions only use the generalized model to perform the tagging process.

ClearNLP utilizes Liblinear L2-regularization, L1-loss support vector classification \cite{fan2008liblinear} for training models and tagging process. It is reported in \cite{choi2012fast} that this method can gain the overall accuracy of 97.46\% for English POS tagging.

\section{Experiments}
\label{sec:exp}
In this section, the process to fix errors in POS-tagged sentences of Vietnamese Treebank corpus is firstly represented. Next, the experimental results of the taggers will be presented.

\subsection{Data processing}

\selectlanguage{vietnamese}
Vietnamese Treebank corpus was built manually. Some serious errors in this data were found while doing experiments. All of those errors are reported in Table~\ref{errors}.

\begin{table*}[t]
\selectlanguage{english}
\footnotesize
\renewcommand{\arraystretch}{1.3}
\caption{Error analysis on Vietnamese Treebank.}
\label{errors}
\selectlanguage{vietnamese}
\begin{center}
\begin{tabular}{l|l|c}
\hline
\multicolumn{1}{c|}{\textbf{Kind of error}}		& \multicolumn{1}{c|}{\textbf{Modification}} 	& \textbf{Occurrence}		\\ \hline
\textit{VN/Np}									& \textit{VN/Ny}      							& 238						\\ \hline
\textit{<number>/M tuổi/Nu}						& \textit{<number>/M tuổi/N}     				& 184						\\ \hline
\begin{tabular}[c]{@{}l@{}}Word segmentation error (two underscores
\\between a pair of syllables)\end{tabular} 	&  Remove one underscore						& 105						\\ \hline
\begin{tabular}[c]{@{}l@{}}Tokenization error (two punctuation marks
\\inside a token)\end{tabular}					& Separate those tokens							& 99						\\ \hline
\icelandic{ð} (Icelandic character)				& đ (Vietnamese character)						& 73     					\\ \hline
More than two tags in one word 					& Remove the wrong tag 							& 50     					\\ \hline
\end{tabular}
\end{center}
\end{table*}

The \#1 row in Table~\ref{errors} presents error in which the word {\it ``VN''} (the abbreviation of {\it ``Việt Nam''}) is tagged as {\it Np} (proper name). The right tag for the word {\it ``VN''} in this case is actually {\it Ny} (abbreviated noun).

The second most frequent error is shown in the \#2 row in Table~\ref{errors}. The context is that a number (tagged with {\it M}) is followed by the word {\it ``tuổi''} ({\it ``years old''}) and the POS tags of {\it ``tuổi''} are not uniform in the whole corpus. There are 184 times in which the tagged sequence is {\it ``<number>/M tuổi/Nu''} ({\it Nu} is unit noun tag which can be used for {\it ``kilograms''}, {\it ``meters''}, etc.) and 246 times that the tagged sequence is {\it ``<number>/M tuổi/N''} ({\it N} is noun). Since the tag {\it N} is more suitable for the word {\it ``tuổi''} in this situation, all 184 occurrences of {\it ``<number>/M tuổi/Nu''} are replaced by the other one.

There are 105 times of word segmentation error in which the separator of syllables is duplicated. Moreover, there are also 99 times of tokenization error, and 73 times that the character \textit{``đ''} is typed wrongly. The last kind of error is that a single word has two POS tags, which happens 50 times.

Obviously, those errors do affect performance of POS taggers significantly. All of them were discovered during the experiments and were fixed manually to improve the accuracy of the taggers.

\selectlanguage{english}

After modifying the corpus, we divide it into ten equal partitions which will be used for 10-fold cross-validation. In each fold, nine of ten partitions are used as the training data, the other one is used as the test set. There are about 1.5\% – 2\% of words in the test set which are unknown in every fold, as shown in Table~\ref{table:dataset}. 

\begin{table}[t]
\footnotesize
\renewcommand{\arraystretch}{1.3}
\caption{The experimental datasets.}
\label{table:dataset}
\begin{center}
\begin{tabular}{c|c|c}
\hline
\textbf{\begin{tabular}[c]{@{}c@{}}Fold\end{tabular}} & \textbf{\begin{tabular}[c]{@{}c@{}}Total number \\of words\end{tabular}} & \textbf{\begin{tabular}[c]{@{}c@{}}Number of\\ unknown words\end{tabular}} \\ \hline
1			& 63277			& 1164				\\ \hline
2			& 63855			& 1203				\\ \hline
3			& 63482			& 1247				\\ \hline
4			& 62228			& 1168				\\ \hline
5			& 59854			& 1056				\\ \hline
6			& 63652			& 1216				\\ \hline
7			& 63759			& 1146				\\ \hline
8			& 63071			& 1224				\\ \hline
9			& 65121			& 1242				\\ \hline
10			& 63552			& 1288				\\ \hline
\end{tabular}
\end{center}
\end{table}

\subsection{Evaluation}

In our experiments, we firstly evaluate the current Vietnamese POS taggers which are vnTagger, JVnTagger and RDRPOSTagger with their default settings. Next, we design a set of features to evaluate the statistical taggers, including two international ones, Stanford Tagger and ClearNLP, and a current Vietnamese one, JVnTagger. There are two terms of the taggers that we measure, which are tagging accuracy and speed. The accuracy is measured using 10-fold cross-validation method on the datasets described above. The speed test is processed on a personal computer with 4 Intel Core i5-3337U CPUs @ 1.80GHz and 6GB of memory. The data used for the speed test is a corpus of 10k sentences collected from Vietnamese websites. This corpus was automatically segmented by UETsegmenter~\cite{nguyen2016hybrid} and contains about 250k words. All taggers use their single-threaded implementation for the speed test. Moreover, the test is processed many times to take the average speed of the taggers. We only use the Java-implemented version of RDRPOSTagger in the experiments because it is claimed by the author that this version runs significantly faster than the other one.

\begin{table*}[t]
\footnotesize
\renewcommand{\arraystretch}{1.3}
\caption{The accuracy results (\%) of current Vietnamese POS taggers with their default settings. \textbf{Ovr.}: the overall accuracy. \textbf{Unk.}: the unknown words accuracy. \textbf{Spd.}: the tagging speed (words per second).}
\label{table:currentresults}
\begin{center}
\begin{tabular}{l|c|c|c|c|c|c|cccccc}
\hline
\multicolumn{1}{c|}{\multirow{3}{*}{\textbf{Feature set}}} & \multicolumn{3}{c|}{\textbf{vnTagger}} & \multicolumn{3}{c|}{\textbf{JVn – MaxEnt}} & \multicolumn{3}{c|}{\textbf{RDRPOSTagger}} & \multicolumn{3}{c}{\textbf{JVn – CRFs}} \\ \cline{2-13} 
\multicolumn{1}{c|}{} & \multicolumn{2}{c|}{Accuracy} & \multirow{2}{*}{Spd.} & \multicolumn{2}{c|}{Accuracy} & \multirow{2}{*}{Spd.} & \multicolumn{2}{c|}{Accuracy} & \multicolumn{1}{c|}{\multirow{2}{*}{Spd.}} & \multicolumn{2}{c|}{Accuracy} & \multicolumn{1}{c}{\multirow{2}{*}{Spd.}} \\ \cline{2-3} \cline{5-6} \cline{8-9} \cline{11-12} \multicolumn{1}{c|}{}

& Ovr.  & Unk. &  & Ovr.  & Unk. &  & \multicolumn{1}{c|}{Ovr.} & \multicolumn{1}{c|}{Unk.} & \multicolumn{1}{c|}{} & \multicolumn{1}{c|}{Ovr.} & \multicolumn{1}{c|}{Unk.} & \multicolumn{1}{c}{} \\ \hline

default & 93.88 & 77.70 & 13k & 93.83 & 79.60 & 50k  & \multicolumn{1}{c|}{93.68} & \multicolumn{1}{c|}{66.07} & \multicolumn{1}{c|}{161k} & \multicolumn{1}{c|}{93.59} & \multicolumn{1}{c|}{69.51} & \multicolumn{1}{c}{47k} \\ \hline

\end{tabular}
\end{center}
\end{table*}

We present the performance of the current Vietnamese taggers in Table~\ref{table:currentresults}. As we can see, the accuracy results of the taggers are pretty similar to each other's with their default feature sets. The most accurate ones are vnTagger and MaxEnt model of JVnTagger. Especially, these two taggers provide very high accuracies for unknown words. Their specialized features for this kind of word seem to be very effective. Inside the JVnTagger toolkit, the two models provides different results. The MaxEnt model of JVnTagger is far more accurate than the CRFs one. Because these two models use the same feature set, we suspect that the MaxEnt model is more efficient than the CRFs one for Vietnamese POS tagging in term of the tagging accuracy. These two models can provide nearly similar tagging speeds which are 50k and 47k words per second. That may be caused by their same feature set (the CRFs model only has an extra feature so its speed is slightly lower). vnTagger has some complicated features such as the conjunction of two tags and uses an outdated version of Stanford Tagger so that its tagging speed is quite low. Meanwhile, the only tagger that does not make use of statistical approach, RDRPOSTagger, produces an impressive tagging speed at 161k words per second. The tagging speed of a transformation-based tagger is mainly based on the speed of its initial tagger. RDRPOSTagger only uses a lexicon for the initial tagger so that it can perform really fast. Nevertheless, its accuracy for unknown words is not good. Its initial tagger just uses some rules to assign initial tags and then it traverses through the rule tree to determine the final result for the each word. Those rules seem to be unable to handle the unknown words well.

\begin{table}[t]
\footnotesize
\renewcommand{\arraystretch}{1.3}
\caption{Feature set designed for experiments of four statistical taggers. \textbf{Dist. Semantics}: distributional semantics, $ds_i$ is the cluster id of the word $w_i$ in the Brown cluster set.}
\label{table:feats}
\begin{center}
\begin{tabular}{l|l}
\hline
\multicolumn{1}{c|}{\textbf{Feature set}} & \multicolumn{1}{c}{\textbf{Template}}							\\ \hline
Simple     		 & \begin{tabular}[c]{@{}l@{}}$w_{\{-2,-1,0,1,2\}}$
										\\ $(w_{-1},w_0), (w_0,w_1),(w_{-1},w_1)$
										\\ $w_0$ has initial uppercase letter?
										\\ $w_0$ contains number(s)?
										\\ $w_0$ contains punctuation mark(s)?
										\\ $w_0$ contains all uppercase letters?
										\\ $w_0$ is first or middle or last token?			\end{tabular}   \\ \hline
Bidirectional    & $(w_0, t_{-1}), (w_0,t_1)$              													\\ \hline
Affix 			 & \begin{tabular}[c]{@{}l@{}}the first syllable of $w_0$
										\\ the last syllable of $w_0$						\end{tabular}   \\ \hline
Dist. Semantics  & $ds_{-1}, ds_0, ds_1$    																\\ \hline
\end{tabular}
\end{center}
\end{table}

The major of the taggers in our experiments is statistical taggers. In the next evaluation, we will create a unique scheme to evaluate these taggers which are Stanford POS Tagger, ClearNLP and JVnTagger. Although vnTagger is also an statistical one, we do not carry it to the second evaluation because it is based on the basis of Stanford Tagger as mentioned.

It is worth repeating that the performance of each statistical tagger is mainly based on its feature set. The feature set we designed for the second evalution is presented in Table~\ref{table:feats}. Firstly, a simple feature set will be applied to all of the taggers. This set only contains the 1-gram, 2-gram features for words and some simple one to catch the word shape and the position of the word in the sentence. Next, we will continuously add more advanced features to the feature set to discover which one makes big impact. The three kinds of avanced feature are bidirectional-context, affix and distributional semantic ones. Whereas, the first and the third one are new to the current Vietnamese POS taggers. The second one is important for predicting the tags of unknown words.

\begin{table*}[t]
\footnotesize
\renewcommand{\arraystretch}{1.3}
\caption{The accuracy results (\%) of the four statistical taggers. \textbf{spl}: the simple feature set. \textbf{bi}: the bidirectional-context feature set. \textbf{affix}: the affix features. \textbf{ds}: the distributional semantic features.}
\label{table:allresults}
\begin{center}
\begin{tabular}{l|c|c|c|c|c|c|cccccc}
\hline
\multicolumn{1}{c|}{\multirow{3}{*}{\textbf{Feature set}}} & \multicolumn{3}{c|}{\textbf{Stanford}} & \multicolumn{3}{c|}{\textbf{ClearNLP}} & \multicolumn{3}{c|}{\textbf{JVn – MaxEnt}} & \multicolumn{3}{c}{\textbf{JVn – CRFs}} \\ \cline{2-13} 
\multicolumn{1}{c|}{} & \multicolumn{2}{c|}{Accuracy} & \multirow{2}{*}{Spd.} & \multicolumn{2}{c|}{Accuracy} & \multirow{2}{*}{Spd.} & \multicolumn{2}{c|}{Accuracy} & \multicolumn{1}{c|}{\multirow{2}{*}{Spd.}} & \multicolumn{2}{c|}{Accuracy} & \multicolumn{1}{c}{\multirow{2}{*}{Spd.}} \\ \cline{2-3} \cline{5-6} \cline{8-9} \cline{11-12} \multicolumn{1}{c|}{}

& Ovr.  & Unk. &  & Ovr.  & Unk. &  & \multicolumn{1}{c|}{Ovr.} & \multicolumn{1}{c|}{Unk.} & \multicolumn{1}{c|}{} & \multicolumn{1}{c|}{Ovr.} & \multicolumn{1}{c|}{Unk.} & \multicolumn{1}{c}{} \\ \hline

spl & 93.96 & 72.19 & 105k & 92.95 & 68.36 & 107k  & \multicolumn{1}{c|}{92.53} & \multicolumn{1}{c|}{67.38} & \multicolumn{1}{c|}{102k} & \multicolumn{1}{c|}{91.57} & \multicolumn{1}{c|}{67.34} & \multicolumn{1}{c}{99k} \\ \hline

spl+bi & 94.24 & 72.40 & 11k  & 93.08 & 68.35 & 93k & \multicolumn{6}{c}{\multirow{3}{*}{N/A}}\\ \cline{1-7}

spl+bi+affix & 94.42 & 78.03 & 10k  & 93.83 & 75.89 & 90k & \multicolumn{6}{c}{}\\ \cline{1-7}

spl+bi+affix+ds & 94.53 & 81.00 & 8k & 94.19 & 79.01 & 64k  & \multicolumn{6}{c}{}\\ \hline

\end{tabular}
\end{center}
\end{table*}

The performance of four statistical taggers are presented in Table~\ref{table:allresults}. Because JVnTagger does not support the bidirectional-context features so we do not have results for it with the feature sets containing this kind of feature. From Table~\ref{table:allresults}, we can see that with the same simple feature set, these taggers can perform with very similar speeds which are approximately 100k words per second. However, their accuracies are different. With the same feature set, the MaxEnt model of Stanford Tagger can significantly outperform the MaxEnt model of JVnTagger. We suspect that it is caused by the algorithm for optimization and some advanced techniques used in Stanford Tagger. Moreover, with this simple feature set, Stanford Tagger also outperforms any other Vietnamese tagger with its default settings in the first evaluation presented above. Stanford Tagger's techniques seem to be really efficient. Next, inside JVnTagger, with the same feature set, the MaxEnt model still performs better than the CRFs one, again, just like the results conducted in Table~\ref{table:currentresults}.

Bidirectional tagging is one of the techniques that have not been applied for current statistical Vietnamese POS taggers. In this experiment, we add two bidirectional-context features which are $(w_0, t_{-1})$ and $(w_0,t_1)$ to the feature set. These two features capture the information of the tags nearby the current word. The results in Table~\ref{table:allresults} reveals that bidirectional-context features help to increase the overall accuracy of Stanford Tagger significantly. Moreover, it also draws the tagging speed of this tagger dramatically. However, this kind of feature only makes small impact for ClearNLP which use SVMs for machine learning process, in terms of tagging accuracy and speed.

Bidirectional-context features do not affect the accuracy of unknown words. Meanwhile, affix feature plays an important role to predict Vietnamese part-of-speech tags. In the next phase of the evaluation, we add the features to catch the first and the last syllable of the current predicting word to discover its impact on the tagging accuracy. As revealed in Table~\ref{table:allresults}, we can conclude that affix features can help to increase the unknown words accuracy sharply, approximately 6\% for both Stanford Tagger and ClearNLP. Especially, those features make a very big improvement in the overall accuracy of ClearNLP.  Moreover, the tagging speeds of these taggers are affected a little bit with these added features.

The last kind of advanced feature is the distributional semantic one. This is a new technique which has been applied to other languages successfully. To extract this feature, we build 1000 clusters of words based on Brown clustering algorithm \cite{Brown:1992:CNG:176313.176316} using a popular implementation\footnote{\url{https://github.com/percyliang/brown-cluster}}. The input corpus consists of 2m articles collected from Vietnamese websites. The result in Table~\ref{table:allresults} shows that distributional semantic features also help to improve the unknown words accuracy of the taggers, at approximately 3\% for both taggers. The overall precision is also increased especially in ClearNLP. The tagging speeds of the tagger are decreased about 20\% to 30\% after adding this kind of feature. 

Overall, Stanford POS Tagger is the one that has the best performance with every feature set. ClearNLP also has a good performance. With the full set of features (spl+bi+affix+ds), both of these two international taggers can outperform the current Vietnamese ones with their default settings in term of tagging accuracy. It leads to the fact that some of the specialized features in current Vietnamese taggers are not really useful. The final results of Stanford Tagger and ClearNLP are also the most accurate ones for Vietnamese POS tagging known to us.

\section{Conclusion}
\label{sec:con}
In this paper, we present an experimental investigation of five part-of-speech taggers for Vietnamese. In the investigation, there are four statistical taggers, Stanford POS Tagger, ClearNLP, vnTagger and JVnTagger. The other one is RDRPOSTagger, a transformation-based tagger. In term of tagging accuracy, we evaluate the statistical taggers by continuously adding several kinds of feature to them. The result reveals that bidirectional tagging algorithm, affix features and distributional semantic features help to improve the tagging accuracy of the statistical taggers significantly. With the full provided feature set, both Stanford Tagger and ClearNLP can outperform the current Vietnamese taggers. In the speed test, RDRPOSTagger produces an impressive tagging speed. The experimental results also show that tagging speed of any statistical tagger is mainly based on its feature set. With a simple feature set, all of the statistical taggers in our experiments can perform at nearly similar speeds. However, giving an complex feature set to the taggers can draw their tagging speeds deeply.


\section*{Acknowledgment}
This work has been supported by Vietnam National University, Hanoi (VNU-H), under Project No. QG.14.04.

\bibliographystyle{IEEEtran}
\bibliography{ref}
\end{document}